\documentclass[a4paper, 10pt, conference]{ieeeconf}      % Use this line for a4
\usepackage{graphicx}
\usepackage{subfig}
\IEEEoverridecommandlockouts                              % This command is only
\title{\LARGE \bf
Enforcing temporal consistency in Deep Learning segmentation of brain MR
images
}

\author{Malav Bateriwala and Pierrick Bourgeat% <-this % stops a space
\thanks{This work was supported by Biomedical Informatics Group, CSIRO}% <-this % stops a space
\thanks{M. Bateriwala is VIBOT student at Heriot-Watt University
        {\tt\small malav.b93@gmail.com}}%
\thanks{P. Bourgeat is senior scientist at Biomedical Informatics Group, CSIRO
        {\tt\small pierrick.bourgeat@csiro.au}}%
}

\begin{document}

\maketitle
\thispagestyle{empty}
\pagestyle{empty}

%%%%%%%%%%%%%%%%%%%%%%%%%%%%%%%%%%%%%%%%%%%%%%%%%%%%%%%%%%%%%%%%%%%%%%%%%%%%%%%%
\begin{abstract}

Longitudinal analysis has great potential to reveal developmental trajectories and monitor disease progression in medical imaging. This process relies on consistent and robust joint 4D segmentation. Traditional techniques are dependent on the similarity of images over time and the use of subject-specific priors to reduce random variation and improve the robustness and sensitivity of the overall longitudinal analysis. This is however slow and computationally intensive as subject-specific templates need to be rebuilt every time. The focus of this work to accelerate this analysis with the use of deep learning. The proposed approach is based on deep CNNs and incorporates semantic segmentation and provides a longitudinal relationship for the same subject. The proposed approach is based on deep CNNs and incorporates semantic segmentation and provides a longitudinal relationship for the same subject. The state of art using 3D patches as inputs to modified Unet provides results around ${0.91 \pm 0.5}$ Dice and using multi-view atlas in CNNs provide around the same results. In this work, different models are explored, each offers better accuracy and fast results while increasing the segmentation quality. These methods are evaluated on 135 scans from the EADC-ADNI Harmonized Hippocampus Protocol. Proposed CNN based segmentation approaches demonstrate how 2D segmentation using prior slices can provide similar results to 3D segmentation while maintaining good continuity in the 3D dimension and improved speed. Just using 2D modified sagittal slices provide us a better Dice and longitudinal analysis for a given subject. For the ADNI dataset, using the simple UNet CNN technique gives us ${0.84 \pm 0.5}$ and while using modified CNN techniques on the same input yields ${0.89 \pm 0.5}$. Rate of atrophy and RMS error are calculated for several test cases using various methods and analyzed. 

\end{abstract}

%%%%%%%%%%%%%%%%%%%%%%%%%%%%%%%%%%%%%%%%%%%%%%%%%%%%%%%%%%%%%%%%%%%%%%%%%%%%%%%%
\section{INTRODUCTION}

Magnetic resonance imaging (MRI) is an important tool used by medical professionals for the diagnosis of patients with neuro and brain related disorders. MRI is usually the modality of choice for structural brain analysis as it is a non-invasive procedure, which provides images with high soft tissues contrast and high spatial resolution. One of the most critical area of the brain in most dementia is the hippocampus. The study of the human hippocampus has gained attention from the neuroscience and neuroimaging communities due to its connection with memory \cite{Nogovitsyn2019Testing} and an array of neurological disorders, especially Alzheimer's disease (AD). \cite{ClaessonDeep}

Existing treatments are only effective in the early phase of the disease, and even then, their effect is highly variable among patients\cite{Akkus2017Deep}. In mild Alzheimer disease, these neurological changes cause loss of neurons in the hippocampus, and these changes result in decreased volume. Many MRI studies have indeed reported smaller volumes in AD patients compared to controls\cite{Bondiau2005Atlasbased}, which indirectly reflects faster rate of atrophy in AD. However, global hippocampus volume is not always sensitive enough to follow changes within a single population, which may reflect conversion from healthy state to disease progression\cite{He2013Hippocampus}. A more accurate way of measuring atrophy is the use of repeated MRI scans of the same individual at certain time interval. This can then be used to monitor the disease progression in patients. More research on longitudinal dataset for volumetric studies and reported a higher rate of hippocampus volume loss in patients with AD subjects compared to healthy individuals\cite{Bondiau2005Atlasbased}. 

\section{RELATED WORK}

2D CNNs neural network architectures are implemented in most cases for application such as image segmentation, prediction and classification. Brain images are represented as 3D volumes in MRI or any other format, but the use of 2D CNNs is motivated because of 2 reasons namely, reduced computational requirements and processing speed. After 2016, Interest in 3D CNNs for brain image segmentation has increased. So with current GPU speed and capabilities , more CNNs work is going on for neuroimaging problems.\cite{Carmo2019Extended} In 2013, a paper looking at temporal analysis of hippocampus volume was introduced using Rate of Regional in Hippocampal Atrophy.\cite{ClaessonDeep} Fig. 1 illustrates the rate of change in size of hippocampus as a function of age. Multivariate analysis of variance was used to assess the effects of age and gender on the metabolite ratios and volume measures.\cite{Frank2013Evaluating}

Many recent initiatives are promoting the development of brain segmentation algorithms such as Unet, which resulted in increase of novel deep learning algorithms and techniques. But all the current method require self tuning to a certain extend for each application and only a handful of CNN-algorithms could operate under reasonable computing requirements. All these novel algorithms have not been tested in independent datasets, just on a common dataset. These segmentation are then post-processed to calculate the rate of change in hippocampal atrophy. Till now, volumetric analysis for temporal consistency in deep learning has not been studied. 

\begin{figure}[h!]
    \centering
    \includegraphics[width=3.5cm]{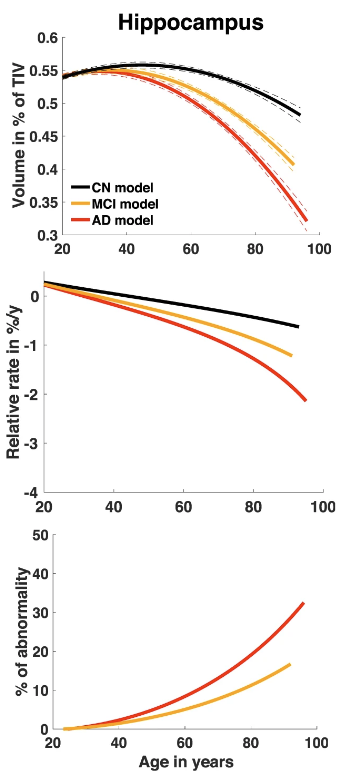}
    \caption {Hippocampus relative volume with respect to function of age \cite{Frank2013Evaluating} }
\end{figure}

Large amount of work in Classification and segmentation of Hippocampus is carried out by a group in China led by Liu M. Their work on 2D dataset with different types of configuration but still there goal is just to classify or segment for a certain subject. As given in the paper "Classification of Alzheimer's Disease by Combination of Convolutional and Recurrent Neural Networks Using FDG-PET Images", 2 techniques are implemented namely UNet and a GRU to segment hippocampus in brain MRI images.\cite{Liu2018Classification} The results obtained are ${0.95 \pm 0.5}$ but results are achieved by cherry picking the data. Another paper was presented by this group for hippocampus segmentation using multi-view method which archived a dice score of $ {0.94 \pm 0.5}$. The implemented model is shown in figure 2.4. In this method they combine segmentation of 3 different views and finally get the final segmentation output using the post processing methods. But here the volumetric analysis of test cases was not carried out. \cite{Nogovitsyn2019Testing}

\section{METHOD}

\subsection{Dataset}
The dataset generated for 2D segmentation training are of 3 types. Both dataset used for training , are created using the same method. For each subject, their 3D MRI and manual is are converted into numpy array. Then fixed number of slices are taken from each scan, as it needs to be consistent and unbiased.
\begin{itemize}
    \item \textbf{Same Slice Data Set} : This dataset is used to compare the results with the proposed method. It is extracted by concatenated the original slice with the same slices. 3 Channel image is extracted instead of 1 Channel. For labels, the slice is extracted and is of 1 channel. Each of these datasets are again divided into 2 parts , one with zero padding by resizing it to 256*256 keeping the original unchanged and the second dataset being the original images  and then re-sized to 128*128.
    \item \textbf{Stacked Data Set} : This dataset is used to train the model in such a way that longitudinal error for different time points is lesser than same slide data set. To extract this dataset images, the original slice is combined with the previous slice and the next slice to make it a 3 channel image. For the label , two different datasets are created. One of the label dataset is the original one slice, extracted without any modifications, and the second one is extracted the same way as the input image and then the 3 Channel label is converted to 1 Channel. Each of these datasets are again divided into 2 parts , one with zero padding by resizing it to 256*256 keeping the original unchanged and the second dataset being the original images  and then re-sized to 128*128.
    \item \textbf{Using CeterCrop augmentation} : This dataset is used for obtaining better segmentation results. It is created of Images size(96*96) from 189*233 dataset. This dataset is generated as a patch of the original images by cropping from the center by size 96*96. This is done in order to extract features of hippocumpus better.
\end{itemize}
\subsection{Models}

\begin{figure}[h!]
    \centering
    \includegraphics[width=8cm]{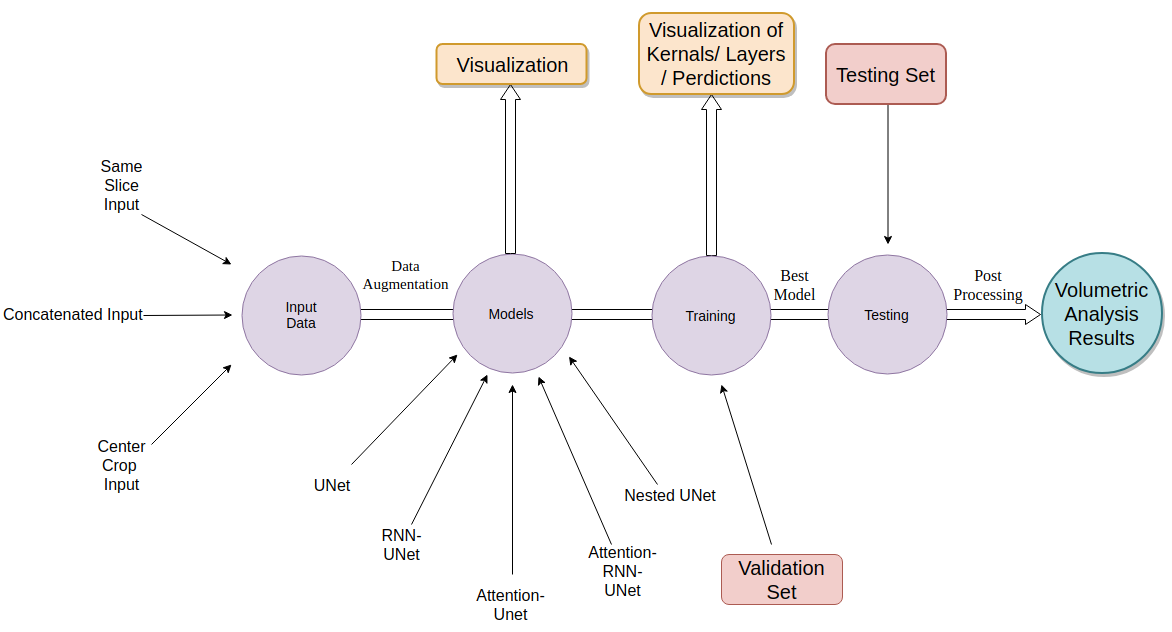}
    \caption {Methodology}
\end{figure}

The main aim is to implement different models and different approaches to see which provides consistency and better volumetric analysis. Till 2017 segmentation was carried out using techniques such a  AlexNet, Unet etc. Each of these methods have unique feature and have different architecture. For medical Imaging, Unet was generally used as it introduced the feature of 'skip connection or architecture'. Skip connections made the training of very deep networks possible and have become an indispensable component in a variety of neural architectures. Skip connection as the name suggests skips some layer in the neural network and feeds the output of one layer as the input to the next layer as well as some other layer (instead of only the next layer). Skip connections help traverse information in deep neural networks. Gradient information can be lost as we pass through many layers, this problem is called as vanishing gradient. The main advantages of skip connection are they pass feature information to lower layers so using it to classify minute details becomes easier and in turn helps to solve the issue of vanishing gradient. During Max-pooling some amount of spatial information is lost and skip connections also can help in that case, and even increases the accuracy of segmentation because final layer feature will have more information.

\begin{itemize} 
    \item \textbf{Unet} : Unet architecture was introduced in the paper "U-Net: Convolutional Networks for Bio-medical Image Segmentation" in March 2015. It was developed by Olaf Ronneberger et al. for Bio-Medical Image Segmentation.{\cite{A2015}} The Unet architecture contains two paths, first path is the contraction path (also called encoder part) which is used to capture the context of the image. The encoder is just a traditional stack of convolutional and max pooling layers. The second path is the symmetric expanding path (also called as the decoder part) which is used to enable precise localization using transposed convolutions. Thus it is an end to end FCN connection but instead this only contains Convolutional layers and does not contain any Dense layer because of which such architecture it can not accept image of any size. 
    
    In this type of architecture, 2 consecutive Convolutions layers are applied. This is done because when pooling is applied, lot of information is lost as the dimensions of the data decreases exponentially. So convolution layers are stacked before each pooling, so it can build up better representations of the data without quickly losing all the spatial information. Also 2 types of basic Unet are defined in the work, one which doesn't use any dropouts and normalization and the other which uses both dropouts and batch normalization. The results of the 2nd method are better and will be discussed in the results section.
    \item\textbf{Attention-Unet} : The attention mechanism enables the neural network to focus on relevant parts of the input data more than the irrelevant parts when performing any kind of operation. It aims to capture image perception as humans and it does that by focusing more on the specific part of the image instead of the complete data.  Attention recreates this mechanism for neural networks. This method was introduces by Ozan Oktay, Jo Schlemper in the 2018 paper "Attention U-Net: Learning Where to Look for the Pancreas". {\cite{Oktay2018Attention}}. While it is mostly used in NLP , it can be used on images where it helps focusing more on a particular section of the images where the label is located, and as a result helps generating better segmentation.The main benefits of using this model architecture are as follows: 
\begin{itemize}
    \item In the skip connections input to the decoder, this method highlights the salient features in the image. Thus it helps the model to train better as it focuses on the required location. It doesn't require more CNN depth to localize the label.
    \item Attention gate filters the neuron activation during the forward pass as well as during the backward pass in such a way that the gradients originating from background regions are down weighted during the backward pass.
\end{itemize}
    \item \textbf{Nested-Unet} : The model architectures previously defined were based on skip connections and they have been state-of-the-art for particular applications. New architecture has been introduced in the paper "UNet++: {A} Nested U-Net Architecture for Medical Image Segmentation" which uses skip connection in a different structure. As seen in the Fig. 3, nested Unet (also know as Unet++) starts with an encoder sub-network or backbone followed by a decoder sub-network. The difference introduced is in the skip connection configuration. In nested unet lot of intermediate connections are added. The bottleneck, remains unaffected, and so does the 1st decoder layer above it, but as the 2nd decoder layer is computed, another parameters are added of the previous encoder layer to the skip connections along with last encoder layer features. Thus the input to the 2nd decoder layer would be 2nd last encoder layer along with the last encoder layer. Moving upward to 3rd decoder layer, the same method is followed up like 2nd decoder layer. The input to the 3rd decoder layer will be 3rd last encoder layer along with the input of the 2nd last layer and the last layer of the encoder. It is kind of a pyramidal scheme. The benefit of this type of network is it can fuse the low level features with the high level features effectively. This method is called deep supervision model. The loss is calculated as the combination of the encoder layers added together as seen in the top right in Fig. 3. The training computation time taken by this model was higher than Attention Unet and Unet.
    \begin{center}
        \begin{figure}[h!]
        \includegraphics[scale=0.55]{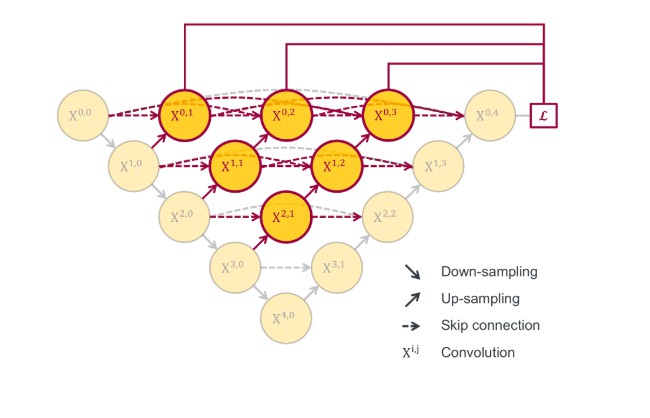}
        \caption{Nested Unet {\cite{Zhou2018UNet}}}
        \centering
    \end{figure}
\end{center}

\end{itemize}

\subsection{Training}

For training the model, different sections were created during programming and they would be called according to the requirement. Dataset was divided into 3 parts, 1st being the training set , 2nd being the validation set and the 3rd being the test set. Also another test set was available, but test set for testing the model was created from the original 135 scans. Each epoch consists of the following steps : \begin{itemize}
    \item Starts with defining all the required parameters for the training such as loss function, optimizer, activation functions, batch size etc. 
    \item Then each batch would be trained on these defined parameters and model created will be used to validate the validation set. The loss function used here are Dice loss and Dice loss + BCE(Binary Cross Entropy) loss weighted loss. Also, weighted loss provides better result in binary case as it focuses more on the label than whole image. The dice is defined as intersection upon union. Dice loss = $$\frac{A \cap B}{A \cup B}$$
    \item After validation step, the model is used to predict a single test image and it is saved to evaluate the model during training.
    \item Then model would be saved along with the visualization of the last layer of the prediction image , also the filter used during  training of the first layer and the gradient flow of the model is saved.
    \item After completion of all the epoch or early stopping, the model is saved with the best parameters and then used for test cases. 
\end{itemize}

\subsection{Evaluation}

After generation of the best model with the provided hyperparameters, it is tested on test cases. Out of the complete 135 scans, we have 5 scans for testing. Here the training , validation and test set all were shuffled randomly. For testing all the images of the test set were passed in the model and a prediction for them was saved. They were then processed with a threshold filter, which removes all the small noise in the prediction. The continuity of the model is checked by saving all the predicted image again in a 3D nifti file. ITK-Snap is used for checking the 3D scan is continuous for all the different views.

The predicted output is then used to calculate the hippocampus volume by summing all the white pixels of the scan. These values are saved in a list for each subject. This process was performed for all the time points for the subject. The list of all the volume of the 10 different time points were plotted and a linear regression was performed for all points. Slope and error is calculated for each different models used and then analyzed.
%%%%%%%%%%%%%%%%%%%%%%%%%%%%%%%%%%%%%%%%%%%%%%%%%%%%%%%%%%%%%%%%%%%%%%%%%%%%%
\section{RESULTS}

\subsection{Segmentation Results}

The dataset trains networks based on image patches extracted around the hippocampus, thus the global spatial information of brain structures are not be perfectly captured. The obtained network models performs well to segment the hippocampal sub-fields around the hippocampus, but cannot recognize those far away from hippocampal region. For example, a patch in the whole image may look similar to the patches in the hippocampus, and will be classified to hippocampal subfields in the testing stage. As a result, there are some isolated false positives outside the hippocampal region. To remove these artifacts automatically, post processing method of thresholding is carried out to remove the pixels which are away from the location of hippocampus. Also thresholding is carried out to remove noise in test cases in all types of datasets and the Dice is calculated. 

Dice score achieved by using 96*96 center cropped images was highest. It achieved score as high as 90.6 Dice with nested Unet model. Using images of size 256*256 gave results around 88.5 Dice on nested Unet model and the least score was given by 128*128 images around 87.3 Dice score. The average Dice Score by using Attention Unet was around 86.5 Dice and for simple Unet it was around 85 Dice. The results comparing to the dataset using the same slices is less, but by a very small margin. Average Dice score for all methods are shown in the table 1 for all types of datasets.
\begin{table}[h!]
\centering
\begin{tabular}{|l|l|l|l|}
\hline
  & Dice & Method         & Data     \\ \hline
1 & 85.4 & Unet           & Same     \\ \hline
2 & 87.7 & Attention-Unet & Same     \\ \hline
3 & 89.2 & Nested Unet    & Same     \\ \hline
4 & 84.4 & Unet           & 3 slices \\ \hline
5 & 86.2 & Attention-Unet & 3 slices \\ \hline
6 & 88.3 & Nested Unet    & 3 slices \\ \hline
\end{tabular}
\caption{Average Dice score for all datasets}
\end{table}

\subsection{Continuity results}

Continuity between slices is the main reason 2D segmentation of same slice was not carried out. The segmentation output was better, but the continuity in the other views was not maintained. Using 3 stacked slices, provides better results.

\section{Temporal consistency Results}

By calculating the volume of brain tissues for each time point, we could check the tissue volume consistency across atlases. The scatter plots fitted with a regression line were shown as results. The volumes of all these brain structures show continuous development along time. Different cases for each brain MRI scan status has been considered for evaluation of the models. 

The plots in Fig. 4 were generated using the cases of healthy patient, MCI patient and AD identified patient. It presents the lifespan models of these structures for these groups. Moreover, relative rate of change is provided. For different techniques the slope of different patients are similar using all models. As in AD patient the volume of the hippocampus is decreasing rapidly in comparison to MCI and healthy patient and it can be visualized in the Fig. 4. The percentage of decrease in volume of AD patient is approximately 3\% and of a healthy patient its around 1.5\%

As shown in the table 2-3-4 the slope for healthy patient is less than that of MCI and AD. It means that more volume of the hipocampus decreases along with time if the cases are of MCI and AD. These was one of the required results to prove that the size of hippocampus decreases over time for MCI and AD patient. The error found for each different models is also mentioned in the all the tables. The error for Healthy cases was least for Nested Unet and highest for Attention Unet. These could again be noticed for MCI and AD cases too. 

\begin{figure}
\centering
\subfloat{\includegraphics[width=5cm]{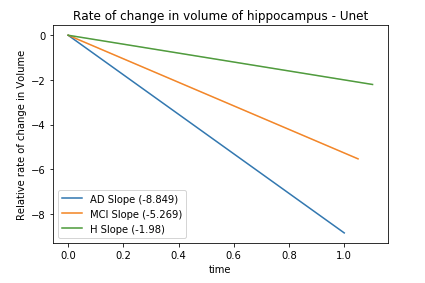}}\hfil
\subfloat{\includegraphics[width=5cm]{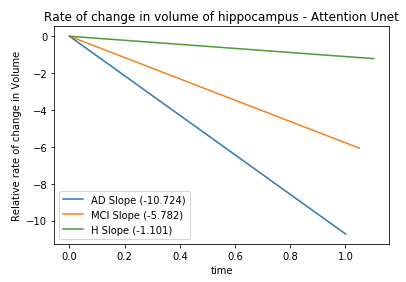}}\hfil 
\subfloat{\includegraphics[width=5cm]{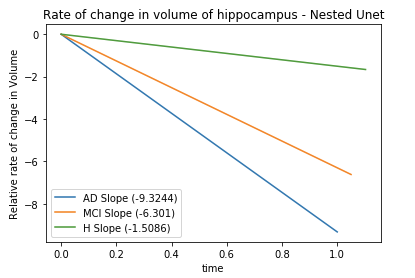}} 
\caption{ Results using dataset size - 256 of same slices}\label{figure}
\end{figure}

\begin{table}[]
\begin{center}
\begin{tabular}{|l|l|l|l|l|}
\hline
  & Method    & Slope (ml/year)           & Error ($mm^3$)     & Status  \\ \hline
a & Unet      & -3.14     & 1204.752  & Healthy \\ \hline
b & Attention & -3.4        & 1361.430  & Healthy \\ \hline
c & Nested    & -2.21      & 1024.334   & Healthy \\ \hline
d & Unet      & -9.39    & 1433.621 & MCI     \\ \hline
e & Attention & -9.7     & 1747.795 & MCI     \\ \hline
f & Nested    & -10.038    & 1347.149 & MCI     \\ \hline
g & Unet      & -8.42    & 3813.653  & AD      \\ \hline
h & Attention & -8.532           & 2811.6316  & AD      \\ \hline
i & Nested    & -9.393    & 3013.885  & AD      \\ \hline
\end{tabular}
\caption {Results of dataset size - 96}
\end{center}
\end{table}

\begin{table}[]
\begin{center}
\begin{tabular}{|l|l|l|l|l|}
\hline
  & Method    & Slope (ml/year)   & Error ($mm^3$)     & Status  \\ \hline
a & Unet      & -1.29  & 388.902   & Healthy \\ \hline
b & Attention & -1.28  & 378.435   & Healthy \\ \hline
c & Nested    & -1.28  & 348.97  & Healthy \\ \hline
d & Unet      & -3.743  & 124.933   & MCI     \\ \hline
e & Attention & -3.508  & 123.933   & MCI     \\ \hline
f & Nested    & -3.18 & 167.419  & MCI     \\ \hline
g & Unet      & -4.44    & 252.511 & AD      \\ \hline
h & Attention & -3.464  & 206.020  & AD      \\ \hline
i & Nested    & -3.108  & 55.844   & AD      \\ \hline
\end{tabular}
\end{center}
\caption {Results of dataset size - 128}
\end{table}

Error for nested Unet for healthy patient, MCI and AD in the table 2 are 732.6103$mm^3$, 1237.69$mm^3$ and 2780.2088$mm^3$ respectively. The results of nested Unet of dataset of size 256 using the longitudinal constraints are 1437.937$mm^3$, 486.125$mm^3$ and 3669.136$mm^3$ respectively. Similar pattern was again observed on comparing other 2 sizes dataset using the similar slices and dataset with longitudinal constraints. 

\begin{table}[]
\begin{center}
\begin{tabular}{|l|l|l|l|l|}
\hline
  & Method    & Slope (ml/year)        & Error ($mm^3$)      & Status  \\ \hline
a & Unet      & -3.86      & 23501.251 & Healthy \\ \hline
b & Attention & -2.36      & 2813.525   & Healthy \\ \hline
c & Nested    & -2.31     & 1437.937  & Healthy \\ \hline
d & Unet      & -6.14      & 654.2790   & MCI     \\ \hline
e & Attention & -7.56        & 2134.7615 & MCI     \\ \hline
f & Nested    & -8.06      & 486.125  & MCI     \\ \hline
g & Unet      & -10.39        & 5561.655  & AD      \\ \hline
h & Attention & -7.99 & 3907.1365  & AD      \\ \hline
i & Nested    & -9.7    & 3669.13650 & AD      \\ \hline
\end{tabular}
\end{center}
\caption {Results of dataset size - 256}
\end{table}

Fig. 5 shows the box of slopes for 5 cases of healthy patient using Nested Unet for all datasets combined. It shows that the difference between them is less for 3 slices datasets than same slices datasets. It shows that accurate results can be obtained using 3 slices for longitudinal analysis and volumetric analysis.  

\begin{figure}
\centering
\subfloat[A]{\includegraphics[width=5cm]{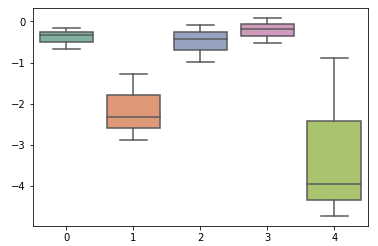}}\hfil 
\subfloat[B]{\includegraphics[width=5cm]{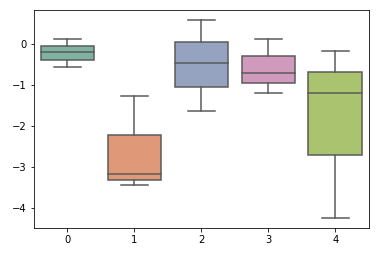}} 
\caption{ Box plot of Slopes of Healthy Patient Cases using Nested Unet : [A] 3 Slice Dataset [B] Same Slices Dataset}\label{figure}
\end{figure}

%%%%%%%%%%%%%%%%%%%%%%%%%%%%%%%%%%%%%%%%%%%%%%%%%%%%%%%%%%%%%%%%%%%%%%%%%%%%%%%%
\section{CONCLUSIONS}

Much of the current work in medical image segmentation is focused on eliminating the need for manual intervention. This work supports the notion that deep neural networks may efficiently and reliably generate hippocampal volumes in both cross-sectional and longitudinal analyses. Here the main contribution is a novel hippocampus segmentation method that achieves nearly the state-of-the-art 2D segmentation performance. Various successful CNN design ideas from the literature were employed in the project. Our modifications to the basic UNet architecture improved DICE by around 3\% and reduced overfitting when testing in other datasets. Not using batch normalization resulted in poor convergence. Training the dataset that is comprised of mostly irregular hippocampus shapes provided better segmentation results. Post processing by thresholding successfully removed noise from small false positive volumes formed during testing. The strategy to use 3 continuous stacked slices didn't increase the score of segmentation, but it did provide a better 3D continuous volume. The method also visually displays generalization in another fairly different dataset using hippocampus as a reference.

A framework for clinical score regression based on segmented hippocampus is proposed. Clinical studies would benefit from a CNN-based fast segmentation algorithm that generates reliable sub-field specific hippocampal volumes. Nested Unet along with deep supervision of features provided the best results among all the different Unet in terms of segmentation results and also for volumetric analysis.The prominent advantages of the method is its not time-consuming and the features generated are better at volumetric analysis for different time points. Also, this method treats all time points the same way and does not make any assumptions on the shape or temporal smoothness of the trajectories. This design increases the flexibility of the proposed Unet segmentation method. As a future work, nvidia vid2vid method will be tested, to further improve the performance of our method. The code is available on GitHub.

%%%%%%%%%%%%%%%%%%%%%%%%%%%%%%%%%%%%%%%%%%%%%%%%%%%%%%%%%%%%%%%%%%%%%%%%%%%%%%%%
\section{ACKNOWLEDGMENTS}

I would like to express our thanks to  CSIRO, Brisbane, where this research has been conducted. I am also thankful to Alzheimer's Disease Neuroimaging Initiative for providing their datasets publicly available to promote development and research.

%%%%%%%%%%%%%%%%%%%%%%%%%%%%%%%%%%%%%%%%%%%%%%%%%%%%%%%%%%%%%%%%%%%%%%%%%%%%%%%%

\bibliographystyle{plain}
\bibliography{refs}
% \begin{thebibliography}{99}
% %\bibliographystyle{plain}
% %\bibliography{refs}

% \bibitem{c1}
% J.G.F. Francis, The QR Transformation I, {\it Comput. J.}, vol. 4, 1961, pp 265-271.

% \bibitem{c2}
% H. Kwakernaak and R. Sivan, {\it Modern Signals and Systems}, Prentice Hall, Englewood Cliffs, NJ; 1991.

% \bibitem{c3}
% D. Boley and R. Maier, "A Parallel QR Algorithm for the Non-Symmetric Eigenvalue Algorithm", {\it in Third SIAM Conference on Applied Linear Algebra}, Madison, WI, 1988, pp. A20.

% \end{thebibliography}

\end{document}